\documentclass[11pt]{article}
\usepackage[margin=1in]{geometry}
\usepackage[T1]{fontenc}
\usepackage[utf8]{inputenc}
\usepackage{amsmath}
\usepackage{amssymb}
\usepackage{graphicx}
\usepackage{booktabs}
\usepackage{float}
\usepackage{caption}
\usepackage{url}
\title{A Prediction-as-Perception Framework for 3D Object Detection}
\author{Song Zhang* ,Haoyu Chen and Ruibo Wang \\Z-one Technology Co., Ltd.\\201 Anyan Road, Jiading, Shanghai, 201805, China\\\texttt{zhang-s03@163.com}, \texttt{chymainbranch@163.com}}
\date{}

\begin{document}
\maketitle

\begin{abstract}
Abstract. Humans combine prediction and perception to observe the world. When faced with rapidly moving birds or insects, we can only perceive them clearly by predicting their next position and focusing our gaze there. Inspired by this, this paper proposes the Prediction-As-Perception (PAP) framework, integrating a prediction-perception architecture into 3D object perception tasks to enhance the model's perceptual accuracy. The PAP framework consists of two main modules: prediction and perception, primarily utilizing continuous frame information as input. Firstly, the prediction module forecasts the potential future positions of ego vehicles and surrounding traffic participants based on the perception results of the current frame. These predicted positions are then passed as queries to the perception module of the subsequent frame. The perceived results are iteratively fed back into the prediction module. We evaluated the PAP structure using the end-to-end model UniAD on the nuScenes dataset. The results demonstrate that the PAP structure improves UniAD's target tracking accuracy by 10\% and increases the inference speed by 15\%. This indicates that such a biomimetic design significantly enhances the efficiency and accuracy of perception models while reducing computational resource consumption.
\end{abstract}


\section{Introduction}

The human brain perceives the world not merely by seeing or hearing and then recognizing, but through a predictive perception framework [ 1, 2 ]. This framework means that the brain constantly generates predictions about future sensory inputs in its environment based on prior knowledge, forming expectations about upcoming events. When actual sensory inputs are received, the brain analyzes the discrepancy between the predictions and the actual inputs, known as prediction errors. The brain then adjusts its internal models to minimize these errors, making both predictions and perceptions more accurate. For example, when we walk on the road, our brain predicts the appearance of vehicles and pedestrians and uses our eyes to verify these predictions, allowing us to safely and quickly avoid these traffic participants. Similarly, when we see insects or birds, we predict their next positions, enabling us to easily track their movements.

However, deep learning technology, which simulates the neural structure and thinking patterns of the human brain, seems to follow a simple input-perception process in 3D object perception. For instance, current state-of-the-art perception and tracking models such as Sparse4 D[ 3 ], HOP[ 4 ], and StreamPETR [ 5 ] perform object detection by inputting image information and applying spatial transformations, attention mechanisms, and other techniques. Similarly, some 3D object trajectory prediction models like THOMAS[ 6 ], Autobot [ 7 ], and GoHome [ 8] follow a single input-perception prediction process.

Inspired by the predictive perception framework of the human brain, we hypothesize that incorporating this framework into 3D object perception or prediction models can effectively enhance the accuracy of both perception and prediction.

\begin{figure}[htbp]
\centering
\includegraphics[width=0.92\linewidth]{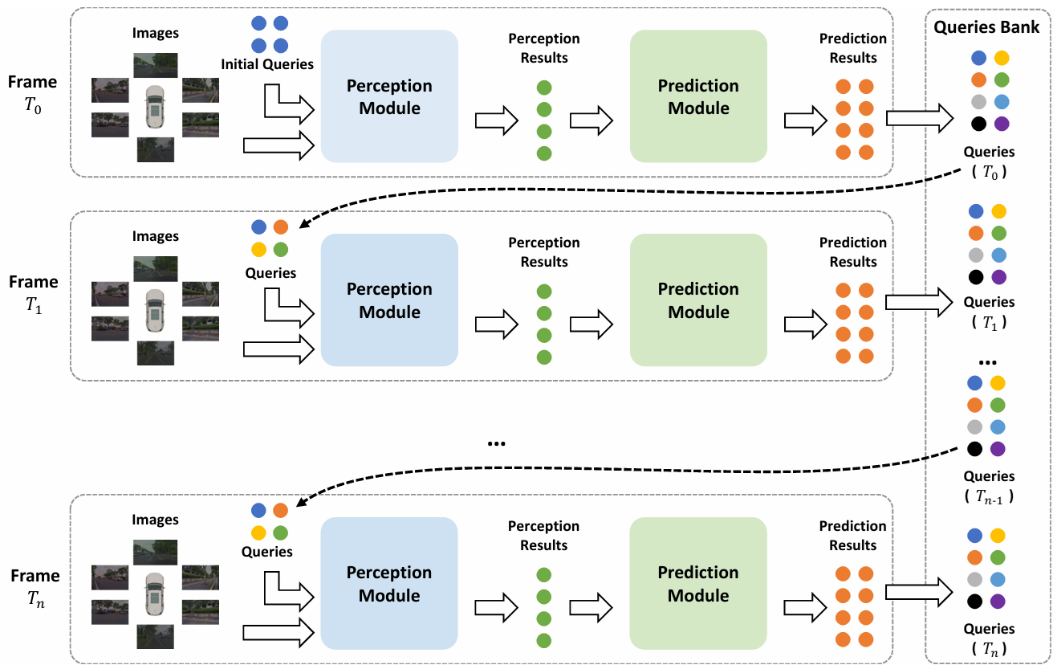}
\caption{Fig. 1. An Overall architecture of PAP. The PAP framework consists of perception and prediction moules. The input to the perception module is the current frame's image and queries, which include both randomly generated queries for the current frame and queries generated from the output results of the prediction module from the previous frame. The input to the pre-diction module is the output results of the perception module, and the output is the possible future positions of traffic participants in the current frame. These position coordinates, once converted into queries, are stored in the queries bank for future frame calls.}
\end{figure}

Therefore, this paper proposes a framework for 3D object perception, tracking, and prediction, named PAP (Prediction As Perception). The PAP framework uses current prediction results as one of the inputs for future perception, thereby simultaneously improving the accuracy of both perception and prediction. PAP consists of two modules: perception and prediction, which communicate via queries. The perception module takes the current frame's image information and the previous frame's prediction queries as input, outputting detection results and their queries. The prediction module takes the current frame's image information and the perception results' queries as input, outputting multi-frame future predictions and corresponding queries, thus iterating over the entire time series data. It should be noted that when the input frame is the first frame of the current data, the historical frame prediction queries input to the perception module will be randomly generated in theory.

We tested UniAD [ 9 ] with the PAP framework on the nuScenes [ 10 ] dataset, and the results showed that UniAD combined with the PAP framework improved the accuracy of perception and prediction tasks by 10\% and 15\%, respectively, de monstrating the effectiveness of our PAP framework.

The main contributions of this paper are as follows:

We propose a predictive perception framework, Prediction as Perception (PAP), for 3D object perception and prediction tasks. This framework uses the prediction results of historical frames as the perception input for the current frame, enhancing both perception and prediction accuracy.

W e integrated the PAP framework with UniAD, resulting in a 10\% improvement in perception performance and a 15\% improvement in prediction performance on the nuScenes dataset.

\section{Related Works}

\subsection{3D Object Detection}

Camera-based 3D object detection is a critical task with many advanced solutions available today. With the rise of multi-sensor setups in autonomous driving, multi-view detection models have become more prevalent. Prominent examples include Bird's Eye View (BEV) detection models that use the Lift-Splat-Shoot (LSS) technique, such as BEVDe t [ 1 1 ] and BEVDepth [ 1 2 ]. These models elevate 2D image features into 3D space through depth estimation and then project them onto the BEV plane for detection with BEV decoders. However, explicit depth estimation often suffers from inaccuracies due to the lack of third-dimensional information. Consequently, a new generation of models has been developed that avoids explicit depth estimation by using transformer architectures. For instance, DETR3 D[ 13 ], Sparse4 D[ 3 ] projects predefined spatial queries onto image features using camera parameters and updates these queries through an attention mechanism, establishing connections between multiple viewpoints. Similarly, models like PETR[ 1 4 ] empl oy positional embeddings, allowing the model to implicitly learn projection relationships without depending on camera parameters, thereby providing an alternative to explicit depth estimation.

\subsection{Future Prediction}

Predicting the future trajectories of traffic participants around the autonomous vehicle is essential for autonomous driving. It provides the basis for the vehicle's path planning and ensures the safety of its driving environment. Initially, trajectory prediction was accomplished using L STM models[ 1 5, 16 ] to forecast the future paths of traffic participants. Later, methods incorporating GANs[ 1 7 ] and other techniques were introduced for trajectory prediction. These classical trajectory prediction approaches typically follow a detect-track-predict pattern[ 18, 1 9 ], where the accumulation of errors between these modules can limit the accuracy of trajectory predictions. Consequently, most current trajectory prediction methods employ end-to-end architectures, which effectively reduce the errors generated during the information exchange between modules. Examples include Fast[ 20 ] and MultiXN et[ 21 ], which uses LiDAR data, and IntentNet [ 22 ], which utilizes HD maps.

\section{Method}

In this section, we elaborate on the main components of the PAP framework. Fig. 1 illustrates the overall structure of the PAP framework. The PAP framework consists of a perception module and a prediction module. First, the current frame's image information and the previous frame's prediction results are input into the perception module. The perception module then outputs the perception results and the corresponding queries, which will be detailed in Section 3.1. The queries from the perception results are fed into the prediction module for future frame prediction. The queries from the prediction results are stored and used as inputs for the perception module in the next frame, which will be discussed in Section 3.2. Finally, in Section 3.3, we describe how we integrate the UniAD model structure into the PAP framework.

\subsection{Perception Module}

The perception module in PAP can directly use existing 3D object perception or tracking models. However, it needs to use models based on attention mechanisms, such as DETR3 D[ 13 ], StreamPETR [ 5 ], and Sparse4 D [ 3 ], which utilize queries for detection. This is because the PAP framework relies on queries for interaction between modules, and the enhancement of the perception module's performance is also achieved through manipulating queries.

In existing attention-based 3D perception or tracking models, the queries for each frame are mostly randomly generated within a certain range and then updated based on the current frame's data. Cross attention is used to establish connections between queries of different frames to achieve detection and tracking. This method of randomly generating queries for each frame can lead to lower object detection efficiency and the loss of temporal cues for the target. The PAP framework incorporates the prediction results of historical frames as queries into the current frame's perception module, explicitly providing the perception module with potential location information of objects. Specifically, the PAP framework replaces some or all of the randomly generated queries in the current frame's perception module with the queries from the prediction results of historical frames, as shown in Fig. 2. This process preserves the temporal cues of the target, making it easier for the perception module to detect and track objects while reducing the computational load associated with updating randomly generated queries.

For example, if the perception module adopts the struct ure of DETR3 D [ 13 ], this process can be simply formulated as follows:

\begin{equation}
c_{i}^{T} = \varnothing^{ref}(q_{i}^{T}), q_{i} \in (q_{random}^{T} \cup q_{predict}^{T-1})
\tag{1}
\end{equation}

where $q_{i}^{T}$ is set of initial queries at time T, $\varnothing^{ref}$ is a nerual network and $c_{i}^{T}$ can be thought of a hypothesis for the center of the i-th box, which is the same as [ 13 ]. In addition, $q_{random}^{T}$ represents randomly generated queries at time T, and $q_{predict}^{T-1}$ represents queries from the prediction module output of the previous frame.

The learning of the newly added prediction result queries is achieved through the combined loss of the perception module and the prediction module. Therefore, the loss of the perception module can remain consistent with that of the original model being used.

\begin{figure}[htbp]
\centering
\includegraphics[width=0.92\linewidth]{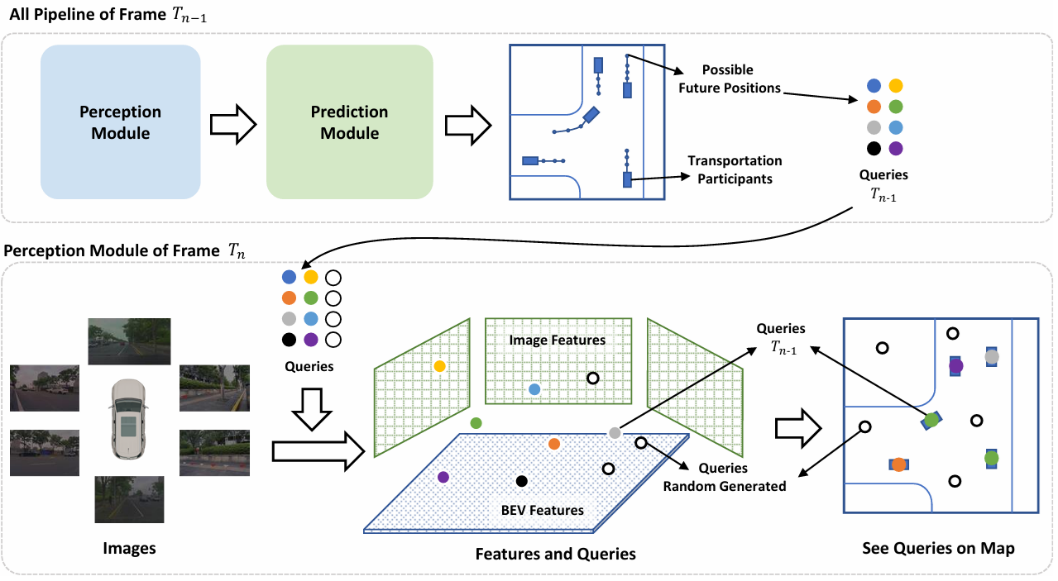}
\caption{Fig. 2. Perception module of PAP. The perception module primarily uses queries to detect objects. Let's assume the perception module has a structure similar to DETR3D[13]. The queries input to the perception module consist of two parts: one part generated from the prediction results of the previous frame, and another part randomly generated for the current frame. When the positions of the queries generated from the previous frame's prediction results are projected onto the map, it is evident that compared to the randomly generated queries, the former are closer to the locations of traffic participants in the current environment.}
\end{figure}

\subsection{Prediction Module}

Since our paper focuses on discussing how incorporating the prediction results from the previous frame into the current frame's perception module can effectively enhance perception accuracy, the prediction module of PAP can directly use the prediction components from existing models. When using existing structures, we only need to embed the prediction results to ensure they match the dimensions of the input queries for the perception module:

\begin{equation}
c_{predict}^{T} = \mathrm{PRED}(\mathrm{PECP}(c_{i}^{T}))
\tag{2}
\end{equation}

\begin{equation}
q_{predict}^{T} = \phi^{embd}(c_{predict}^{T})
\tag{3}
\end{equation}

where $c_{i}^{T}$ is the center of the i-th box at time T, $\mathrm{PECP}(\cdot)$ is the perception module of PAP, and $\mathrm{PRED}(\cdot)$ is the prediction module. $c_{predict}^{T}$ represents the output results of the prediction module, containing the coordinates of the possible future positions of the objects within the scene, $\phi^{embd}$ represents the embedding layer, which can embed the coordinates into queries of a specific dimension. Then, all the queries will be stored with time as the index, facilitating subsequent calls by the perception module.

\begin{figure}[htbp]
\centering
\includegraphics[width=0.92\linewidth]{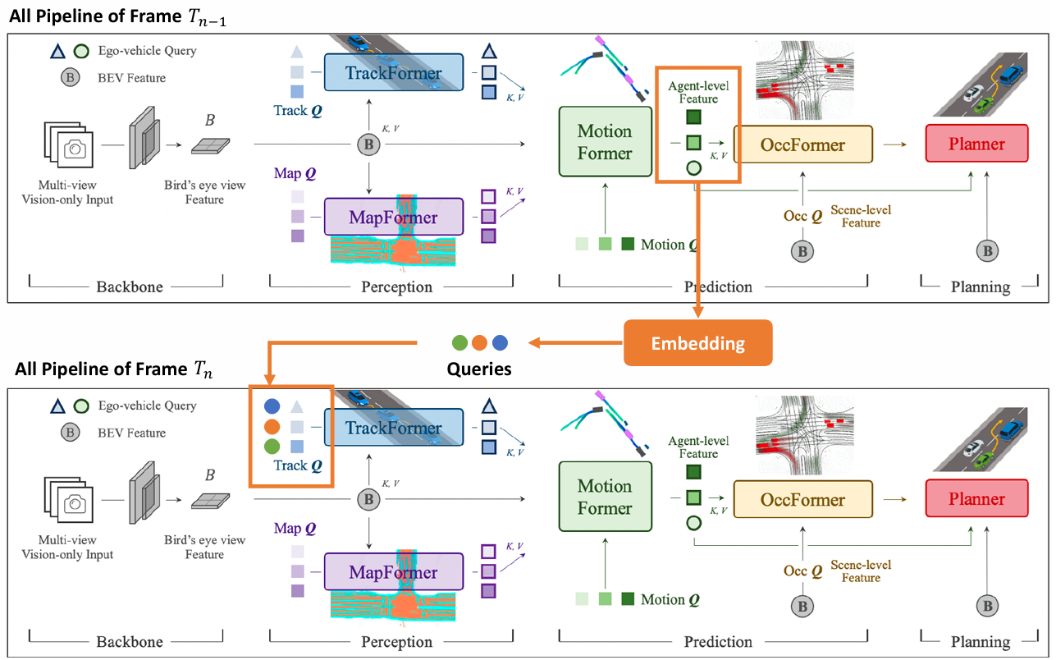}
\caption{Fig. 3. PAP with UniAD[9]. In UniAD, the interaction between modules is inherently based on queries. Therefore, we only need to take the queries output by the Motion Former module in UniAD, embed them to match the dimensions of the Track Queries, and then feed them together with the Track Queries into Track Former for detection.}
\end{figure}

\subsection{PAP with UniAD}

UniAD [ 9 ] is an end-to-end model capable of implementing planning, and it includes perception, prediction, and planning modules, with interactions between these modules relying on queries. Therefore, for experimental efficiency, we directly integrate and validate the PAP framework within the UniAD structure. Since UniAD's perception module uses queries for object detection and the prediction module directly outputs queries of prediction results, we store the prediction results output by UniAD for each frame and replace part of the randomly generated queries in the next frame's perception module with the queries from the previous frame's prediction results, as shown in Fig. 3. The rest of the UniAD structure, such as the planning module and the loss components, remain unchanged.

\section{Experiments}

\subsection{Dataset}

In this study, we tested our method on the nuScenes dataset. The nuScenes dataset consists of 1000 sequences, each sequence is 20 seconds long, with up to 20 frames captured per second. Each sample in the dataset includes images from 6 cameras (front, front-left, front-right, back, back-left, and back-right), along with the extrinsic and intrinsic parameters of all sensors. The nuScenes dataset also provides ground truth information and evaluation methods for 3D object detection, 3D object tracking, and trajectory prediction. The model UniAD, which we used to test the PAP framework, employs the AMOTA (average multi-object tracking accuracy) and AMOTP (average multi-object tracking precision) metrics provided by nuScenes for evaluation in the perception part. We use these two metrics to evaluate the effectiveness of PAP as well.

\subsection{Training \& Inference}

Our experimental setup includes a server with 4 A100 GPUs, a 64-core CPU, and 256GB of RAM. During the experiments, all training parameters and hyperparameters of UniAD were kept consistent with the original model to ensure the fairness of performance comparison after integrating PAP. The total training time was 3 days, and the inference speed was 16 frames per second.

\subsection{Results}

The final experimental results are shown in Table 1 and Table 2. As can be seen, after using the PAP framework, various metrics of UniAD have been optimized. Specifically, AMOTA improved by 0.04, which is a 10\% increase, AMOTP decreased by 0.1, Recall improved by 0.03, and IDS decreased to 826. Additionally, the training time of the original model in our experimental environment was 4 days (91 hours), with an inference speed of 14 frames per second. After integrating the PAP structure, the training time of UniAD was reduced to 3 days (78 hours), and the inference speed increased by 2 frames per second, resulting in a 15\% improvement in efficiency. This demonstrates the effectiveness of the PAP framework.

\begin{table}[htbp]
\centering
\small
\caption{Table 1. Performance comparison between UniAD and UniAD+PAP on the nuScenes validation dataset. UniAD+PAP outperforms UniAD in all performance metrics, and it is also more efficient in terms of training and computation.}
\begin{tabular}{lcccccc}
\toprule
 & AMOTA & AMOTP & Recall & IDS & Training time & FPS \\
\midrule
UniAD & 0.359 & 1.32 & 0.467 & 906 & 91h & 14 \\
UniAD+PAP & 0.395 & 1.22 & 0.493 & 826 & 78h & 16 \\
\bottomrule
\end{tabular}
\end{table}

\begin{table}[htbp]
\centering
\small
\caption{Table 2. Performance metrics of UniAD + PAP for each category.}
\begin{tabular}{lcccc}
\toprule
 & AMOTA & AMOTP & Recall & IDS \\
\midrule
Bicycle & 0.372 & 1.297 & 0.453 & 15 \\
Bus & 0.465 & 1.225 & 0.535 & 8 \\
Car & 0.613 & 0.744 & 0.667 & 405 \\
Motor & 0.438 & 1.253 & 0.5 & 24 \\
Pedest & 0.411 & 1.192 & 0.487 & 342 \\
Trailer & 0.33 & 1.551 & 0.201 & 4 \\
truck & 0.411 & 1.267 & 0.611 & 28 \\
\bottomrule
\end{tabular}
\end{table}

\subsection{Limitations}

Our PAP framework strongly depends on the performance of the original perception and prediction models. Due to time constraints and experimental convenience, we directly validated our idea on UniAD in this study. However, UniAD itself focuses on end-to-end planning, and therefore, its perception and prediction modules individually are not state-of-the-art. In future work, we plan to integrate more advanced perception and prediction models into our PAP framework to achieve better results, and furthermore, we will also conduct ablation studies.

\section{Conclusion}

In this paper, we propose a framework inspired by the functioning of the human brain, called the Prediction As Perception (PAP) framework, aimed at enhancing the accuracy of 3D object perception tasks. The purpose of the PAP framework is to incorporate historical prediction results into the current perception module to improve the model's perception accuracy. The PAP framework consists of perception and prediction modules, which can be directly constructed using existing perception and prediction models. We conducted a preliminary validation of the PAP framework's effectiveness on UniAD, and the results show that the integration of the PAP framework can improve UniAD's object tracking accuracy by 10\%, while also enhancing its training and inference speed by 15\%. This demonstrates the effectiveness of our PAP framework.


\begin{thebibliography}{99}
\bibitem{ref1} A. Bubic, D. Y. Von Cramon, and R. I. Schubotz. Prediction, cognition and the brain. Frontiers in human neuroscience, 4:1094, 2010.
\bibitem{ref2} J. Hohwy. The predictive processing hypothesis. The Oxford handbook of 4E cognition, pages 129-145, 2018.
\bibitem{ref3} X. Lin, T. Lin, Z. Pei, L. Huang, and Z. Su. Sparse4d: Multi-view 3d object detection with sparse spatial-temporal fusion. arXiv preprint arXiv:2211.10581, 2022.
\bibitem{ref4} Z. Zong, D. Jiang, G. Song, Z. Xue, J. Su, H. Li, and Y. Liu. Temporal enhanced training of multi-view 3d object detector via historical object prediction. In Proceedings of the IEEE/CVF International Conference on Computer Vision, pages 3781-3790, 2023.
\bibitem{ref5} S. Wang, Y. Liu, T. Wang, Y. Li, and X. Zhang. Exploring object-centric temporal modeling for efficient multi-view 3d object detection. In Proceedings of the IEEE/CVF International Conference on Computer Vision, pages 3621-3631, 2023
\bibitem{ref6} T. Gilles, S. Sabatini, D. Tsishkou, B. Stanciulescu, and F. Moutarde. Thomas: Trajectory heatmap output with learned multi-agent sampling. arXiv preprint arXiv:2110.06607, 2021.Author, F., Author, S.: Title of a proceedings paper. In: Editor, F., Editor, S. (eds.) CONFERENCE 2016, LNCS, vol. 9999, pp. 1-13. Springer, Heidelberg (2016).
\bibitem{ref7} R. Girgis, F. Golemo, F. Codevilla, M. Weiss, J. A. D'Souza, S. E. Kahou, F. Heide, and C. Pal. Latent variable sequential set transformers for joint multi-agent motion prediction. arXiv preprint arXiv:2104.00563, 2021.
\bibitem{ref8} T. Gilles, S. Sabatini, D. Tsishkou, B. Stanciulescu, and F. Moutarde. Gohome: Graph-oriented heatmap output for future motion estimation. In 2022 international conference on robotics and automation (ICRA), pages 9107-9114. IEEE, 2022.
\bibitem{ref9} Y. Hu, J. Yang, L. Chen, K. Li, C. Sima, X. Zhu, S. Chai, S. Du, T. Lin,W.Wang, et al. Planning oriented autonomous driving. In Proceedings of the IEEE/CVF Conference on Computer Vision and Pattern Recognition, pages 17853-17862, 2023.
\bibitem{ref10} Caesar H , Bankiti V , Lang A H ,et al. nuScenes: A multimodal dataset for autonomous driving[J]. 2019.DOI:10.1109/CVPR42600.2020.01164.
\bibitem{ref11} J. Huang, G. Huang, Z. Zhu, Y. Ye, and D. Du. Bevdet: High-performance multi-camera 3d object detection in bird-eye-view, 2022.
\bibitem{ref12} Y. Li, Z. Ge, G. 223 Yu, J. Yang, Z. Wang, Y. Shi, J. Sun, and Z. Li. Bevdepth: Acquisition of reliable depth for multi-view 3d object detection, 2022.
\bibitem{ref13} Y.Wang, V. Guizilini, T. Zhang, Y.Wang, H. Zhao, and J. Solomon. Detr3d: 3d object detection from multi-view images via 3d-to-2d queries, 2021.
\bibitem{ref14} Y. Liu, T. Wang, X. Zhang, and J. Sun. Petr: Position embedding transformation for multi-view 3d object detection, 2022.
\bibitem{ref15} A. Alahi, K. Goel, V. Ramanathan, A. Robicquet, L. Fei-Fei, and S. Savarese. Social lstm: Human trajectory prediction in crowded spaces. In Proceedings of the IEEE conference on computer vision and pattern recognition, pages 961-971, 2016.Author, F.: Article title. Journal 2(5), 99-110 (2016).
\bibitem{ref16} N. Deo and M. M. Trivedi. Convolutional social pooling for vehicle trajectory prediction. In Proceedings of the IEEE conference on computer vision and pattern recognition workshops, pages 1468-1476, 2018.
\bibitem{ref17} A. Gupta, J. Johnson, L. Fei-Fei, S. Savarese, and A. Alahi. Social gan: Socially acceptable trajectories with generative adversarial networks. In Proceedings of the IEEE conference on computer vision and pattern recognition, pages 2255-2264, 2018.
\bibitem{ref18} Y. Chai, B. Sapp, M. Bansal, and D. Anguelov. Multipath: Multiple probabilistic anchor trajectory hypotheses for behavior prediction, 2019.
\bibitem{ref19} J. Hong, B. Sapp, and J. Philbin. Rules of the road: Predicting driving behavior with a convolutional model of semantic interactions, 2019.
\bibitem{ref20} W. Luo, B. Yang, and R. Urtasun. Fast and furious: Real time end-to-end 3d detection, tracking and motion forecasting with a single convolutional net, 2020.
\bibitem{ref21} N. Djuric, H. Cui, Z. Su, S. Wu, H. Wang, F.-C. Chou, L. S. Martin, S. Feng, R. Hu, Y. Xu, A. Dayan, S. Zhang, B. C. Becker, G. P. Meyer, C. Vallespi-Gonzalez, and C. K. Wellington. Multixnet: Multiclass multistage multimodal motion prediction, 2021.
\bibitem{ref22} S. Casas, W. Luo, and R. Urtasun. Intentnet: Learning to predict intention from raw sensor data, 2021.
\end{thebibliography}
\end{document}